\documentclass[letterpaper]{article} 
\usepackage{aaai23}  
\usepackage{times}  
\usepackage{helvet}  
\usepackage{courier}  
\usepackage[hyphens]{url}  
\usepackage{graphicx} 
\usepackage{natbib}  
\usepackage{caption} 
\usepackage{algorithm}
\usepackage{algorithmic}

\newcommand{\xadv}{x_{\textnormal{adv}}}
\newcommand{\yhat}{\hat{y}}
\newcommand{\xtest}{x_{\textnormal{test}}}
\newcommand{\ytest}{y_{\textnormal{test}}}

\newcommand{\loss}{\textsf{loss}}
\newcommand{\logit}{\textsf{logit}}

\newcommand{\Dtrain}{D_{\textnormal{train}}}
\newcommand{\Dtest}{D_{\textnormal{test}}}
\newcommand{\Dpoison}{D_{\textnormal{poison}}}

\newcommand{\cS}{\mathcal{S}}

\newcommand{\cT}{\mathcal{T}}

\newcommand{\bbR}{\mathbb{R}}

\usepackage{amsmath,amsthm,amssymb}
\usepackage{booktabs}
\usepackage{adjustbox}
\usepackage{newfloat}
\usepackage{listings}
\DeclareCaptionStyle{ruled}{labelfont=normalfont,labelsep=colon,strut=off} 
\floatstyle{ruled}
\newfloat{listing}{tb}{lst}{}
\floatname{listing}{Listing}
\nocopyright 

\title{Holistic Adversarial Robustness of Deep Learning Models}
\author{
    Pin-Yu Chen\textsuperscript{\rm 1}, 
    Sijia Liu\textsuperscript{\rm 2}
    
}
\affiliations{
    \textsuperscript{\rm 1} IBM Research\\
    \textsuperscript{\rm 2} Michigan State University\\

    pin-yu.chen@ibm.com, liusiji5@msu.edu
}

\usepackage{bibentry}

\begin{document}

\maketitle

\begin{abstract}
Adversarial robustness studies the worst-case performance of a machine learning model to ensure safety and reliability. With the proliferation of deep-learning-based technology, the potential risks associated with model development and deployment can be amplified and become dreadful vulnerabilities. This paper provides a comprehensive overview of research topics and foundational principles of research methods for adversarial robustness of deep learning models, including attacks, defenses, verification, and novel applications.
\end{abstract}

\section{Introduction}
Deep learning \cite{lecun2015deep} is a core engine that drives recent advances in artificial intelligence (AI) and machine learning (ML), and it has broad impacts on our society and technology. However, there is a growing gap between AI technology's creation and its deployment in the wild. One critical example is the lack of \textit{robustness}, including natural robustness  to data distribution shifts, ability in generalization and adaptation to new tasks, and worst-case robustness when facing an adversary (also known as \textit{adversarial robustness}).
According to a recent Gartner report\footnote{\url{https://www.gartner.com/smarterwithgartner/gartner-top-10-strategic-technology-trends-for-2020}},
30\% of cyberattacks by 2022 will involve data poisoning, model theft or adversarial examples. 
However, the industry seems underprepared. In a survey of 28 organizations spanning small as well as large organizations, 25 organizations did not know how to secure their AI/ML systems \cite{kumar2020adversarial}.
Moreover, various practical vulnerabilities and incidences incurred by AI-empowered systems have been reported in real life, such as Adversarial ML Threat Matrix\footnote{\url{https://github.com/mitre/advmlthreatmatrix}} and AI Incident Database\footnote{\url{https://incidentdatabase.ai/}}.

To prepare deep-learning enabled AI systems for the real world and to familiarize researchers with the error-prone risks hidden in the lifecycle of AI model development and deployment -- spanning from data collection and processing, model selection and training, to model deployment and system integration --
this paper aims to provide a holistic overview of adversarial robustness for deep learning models. The research themes include (i) attack (risk identification and demonstration), (ii) defense (threat detection and mitigation), (iii) verification (robustness certificate), and (iv) novel applications. 
In each theme, the fundamental concepts and key research principles will be presented in a unified and organized manner. 
This paper takes an overarching and holistic approach to introduce adversarial robustness of deep learning models based on the terminology of an AI lifecycle in development and deployment, which differs from existing survey papers that provide an in-depth discussion on a specific threat model. The main goal of this paper is to deliver a primer that provides basic concepts, systematic knowledge, and categorization of this rapidly evolving research field to the general audience and the broad AI/ML research community.

\begin{figure}
    \centering
    \includegraphics[width=0.99\columnwidth]{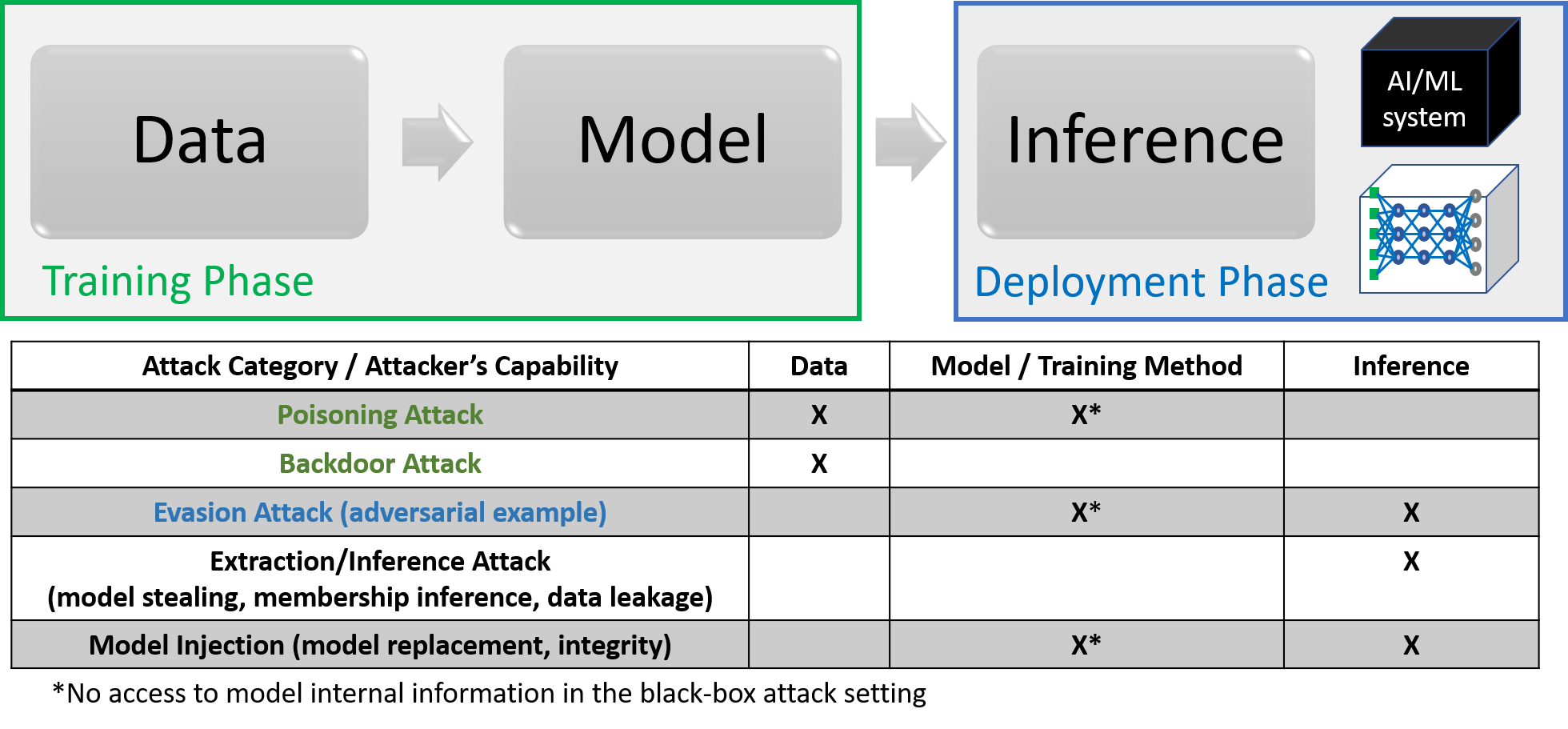}
    \caption{Holistic view of adversarial attack categories and capabilities (threat models) in the training and deployment phases. The three types of attacks highlighted in colors (poisoning/backdoor/evasion attack) are the major focus of this paper. In the deployment phase, the target (victim) can be an access-limited black-box system (e.g. a prediction API) or a transparent white-box model.  }
    \label{Fig_holistic}
\end{figure}

\begin{table*}[t]
\centering
\begin{tabular}{@{}l|l@{}}
\toprule
Symbol                                                                                                    & Meaning                                                                   \\ \midrule
$f_{\theta}: \bbR^{d} \mapsto [0,1]^K$      & $K$-way neural network classification model parameterized by $\theta$                                             \\
$\logit:\bbR^d \mapsto \bbR^K$         & logit (pre-softmax) representation                        \\
$(x,y)$                                                                             & data sample $x$ and its associated groundtruth label $y$                                \\
$\yhat_{\theta}(x) \in [K]$ &  top-1 label prediction of $x$ by $f_{\theta}$                                          \\
$\xadv $          & adversarial example of $x$                  \\
$\delta $          & adversarial perturbation to $x$ for evasion attack               \\
$\Delta $          & universal trigger pattern for backdoor attack                  \\
$t \in [K]$          & target label for targeted attack                             \\
$\loss(f_{\theta}(x),y)$          & classification loss (e.g. cross entropy)                            \\
$g$          & attcker's loss function                           \\
$\Dtrain$ / $\Dtest$       &     original training / testing dataset               \\
 $\cT$      & data transformation function                 \\
\bottomrule
\end{tabular}
\caption{Mathematical notation. }
\label{tab:notation}
\end{table*}

\begin{table*}[t]
\centering
\begin{tabular}{@{}l|l@{}}
\toprule
Attack                                                                                              & Objective                                                                  \\ \midrule
Poisoning  &  Design a poisoned dataset $\Dpoison$ such that models trained                                       \\ 
                                                             &   on $\Dpoison$  fail to generalize on $\Dtest$ (i.e. $\yhat_\theta(\xtest)\neq \ytest$)                     \\
Backdoor                                                                         & Embed a trigger $\Delta$ with a target label $t$ to $\Dtrain$ such that                     \\
                                                                      &     $\yhat_\theta(\xtest)=\ytest$ but $\yhat_\theta(\xtest+\Delta) = t$                        \\
Evasion (untargeted)          & Given  $f_\theta$, find $\xadv$ such that $\xadv$ is similar to $x$ but   $\yhat_\theta(\xadv) \neq y$               \\
Evasion (targeted)          & Given  $f_\theta$, find $\xadv$ such that $\xadv$ is similar to $x$ but   $\yhat_\theta(\xadv) =t$                          \\
\bottomrule
\end{tabular}
\caption{Objectives of adversarial attacks. }
\label{tab:objective}
\end{table*}

Figure \ref{Fig_holistic} shows the lifecycle of AI development and deployment and different adversarial threats corresponding to attackers' capabilities (also known as threat models). The lifecycle is further divided into two phases.
The \textit{training} phase includes data collection and pre-processing, as well as model selection (e.g. architecture search and design), hyperparameter tuning, model parameter optimization, and validation. After model training, the model is ``frozen'' (fixed model architecture and parameters) and is ready for deployment. Before deployment, there are possibly some post-hoc model adjustment steps such as model compression and quantification for memory/energy reduction, calibration or risk mitigation.
The frozen model providing inference/prediction can be deployed in a white-box or black-box manner. The former means the model details are transparent to a user (e.g. releasing the model architecture and pre-trained weights for neural networks), while the latter means a user can access model predictions but does not know what the model is (i.e., an access-limited model), such as a prediction API. The gray-box setting is a mediocre scenario that assumes a user knows partial information about the deployed model.
In some cases, a user may have knowledge of the training data and the deployed model is black-box, such as in the case of  an AI automation service that only returns a model prediction portal based on user-provided training data. We also note that these two phases can be recurrent: a deployed model can re-enter the training phase with continuous model/data updates.

Throughout this paper, we focus on adversarial robustness of neural networks for classification tasks. Many principles in classification can be naturally extended to other machine learning tasks, which will be discussed in Section \ref{sec_remark}.
Based on Figure \ref{Fig_holistic}, this paper will focus on training-phase and deployment-phase attacks driven by the limitation of current ML techniques. While other adversarial threats concerning model/data privacy and integrity are also crucial, 
such as model stealing, membership inference, data leakage, and model injection, which will not be covered in this paper. 
We also note that adversarial robustness of non-deep-learning models such as support vector machines has been investigated. 
We refer the readers to \cite{biggio2018wild} for the research evolution in adversarial machine learning.

Table \ref{tab:notation} summarizes the main mathematical notations. We use $[K]=\{1,2,\ldots,K\}$ to denote the set of $K$ class labels. 
Without loss of generality, we assume the data inputs are vectorized (flattened) as $d$-dimensional real-valued vectors, and the output (class confidence) of the $K$-way neural network classifier $f_\theta$ is nonnegative and sum to $1$ (e.g. softmax as the final layer), that is, $\sum_{k=1}^K [f_\theta(\cdot)]_k=1$. The adversarial robustness of real-valued continuous data modalities such as image, audio, time series, and tabular data can be studied based on a unified methodology. For discrete data modalities such as texts and graphs, one can leverage their real-valued embeddings \cite{lei2018discrete}, latent representations, or continuous relaxation of the problem formulation (e.g. topology attack in terms of edge addition/deletion in graph neural networks \cite{xu2019topology}). Unless specified, in what follows we will not further distinguish data modalities.

\section{Attacks}

This section will cover mainstream adversarial threats that aim to manipulate the prediction and decision-making of an AI model through training-phase or deployment-phase attacks. Table \ref{tab:objective} summarizes their attack objectives.

\subsection{Training-Phase Attacks}

Training-phase attacks assume the ability to modify the training data to achieve malicious attempts on the resulting model, which can be realized through noisy data collection such as crowdsourcing. Specifically, the memorization effect of deep learning models \cite{zhang2016understanding,carlini2019secret} can be leveraged as vulnerabilities. We note that sometimes the term ``data poisoning'' entails both poisoning and backdoor attacks, though their attack objectives are different. 

\paragraph{Poisoning attack} aims to  design a poisoned dataset $\Dpoison$ such that models trained on $\Dpoison$ will fail to generalize on $\Dtest$ (i.e. $\yhat_\theta(\xtest)\neq \ytest$) while the training loss remains similar to clean data. The poisoned dataset $\Dpoison$ can be created by modifying the original training dataset $\Dtrain$, such as label flipping, data addition/deletion, and feature modification. The rationale is that training on $\Dpoison$ will land on a ``bad'' local minimum of model parameters.

To control the amount of data modification and reduce the overall accuracy on $\Dtest$ (i.e. test accuracy), poisoning attack often assumes the knowledge of the target model and its training method \cite{jagielski2018manipulating}. \cite{liu2020min} proposes black-box poisoning with additional conditions on the training loss function. Targeted poisoning attack aims at manipulating the prediction of a subset of data samples in $\Dtest$, which can be accomplished by clean-label poisoning (small perturbations to a subset of $\Dtrain$ while keeping their labels intact) \cite{shafahi2018poison,zhu2019transferable} or gradient-matching poisoning \cite{geiping2020witches}.

\paragraph{Backdoor attack} is also known as Trojan attack. The central idea is to embed a universal trigger $\Delta$ to a subset of data samples in $\Dtrain$ with a modified target label $t$ \cite{BadNet_Access}. Examples of trigger patterns are a small patch in images and a specific text string in sentences. Typically, backdoor attack only assumes access to the training data and does not assume the knowledge of the model and its training. The model $f_\theta$ trained on the tampered data is called a backdoored (Trojan) model.
Its attack objective has two folds: (i) High standard accuracy in the absence of trigger -- the backdoored model should behave like a normal model (same model trained on untampered data), i.e., $\yhat_\theta(\xtest)=\ytest$. (ii) High attack success rate in the presence of trigger -- the backdoored model will predict any data input with the trigger as the target label $t$, i.e., $\yhat_\theta(\xtest+\Delta)=t$.
Therefore, backdoor attack is stealthy and insidious. 
The trigger pattern can also be made input-aware and dynamic \cite{nguyen2020input}.

There is a growing concern of backdoor attacks in emerging machine learning systems featuring collaborative model training with local private data, such as federated learning \cite{bhagoji2019analyzing,bagdasaryan2020backdoor}. Backdoor attacks can be made more insidious by leveraging the innate local model/data heterogeneity and diversity \cite{zawad2021curse}.  \cite{xie2020dba} proposes distributed backdoor attacks by trigger pattern decomposition among malicious clients to 
make the attack more stealthy  and effective. 
We also refer the readers to the detailed survey of these two attacks in \cite{goldblum2020data}.

\begin{table}[t]
\centering
\begin{tabular}{@{}l|l@{}}
\toprule
Norm                                                                                             & Meaning                                                                  \\ \midrule
$\ell_0$         &   number of modified features                             \\ 
$\ell_1$         &   total changes in modified features                          \\ 
$\ell_2$         &   Euclidean distance between $x$ and $\xadv$
\\
$\ell_\infty$         &  maximal change in modified features
\\
\bottomrule
\end{tabular}
\caption{$\ell_p$ norm similarity measures for additive perturbation $\delta=\xadv-x$. The change in each feature (dimension) between $x$ and $\xadv$ is measured in absolute value.}
\label{tab:similarity}
\end{table}

\begin{figure*}[t]
    \centering
    \includegraphics[width=0.97\textwidth]{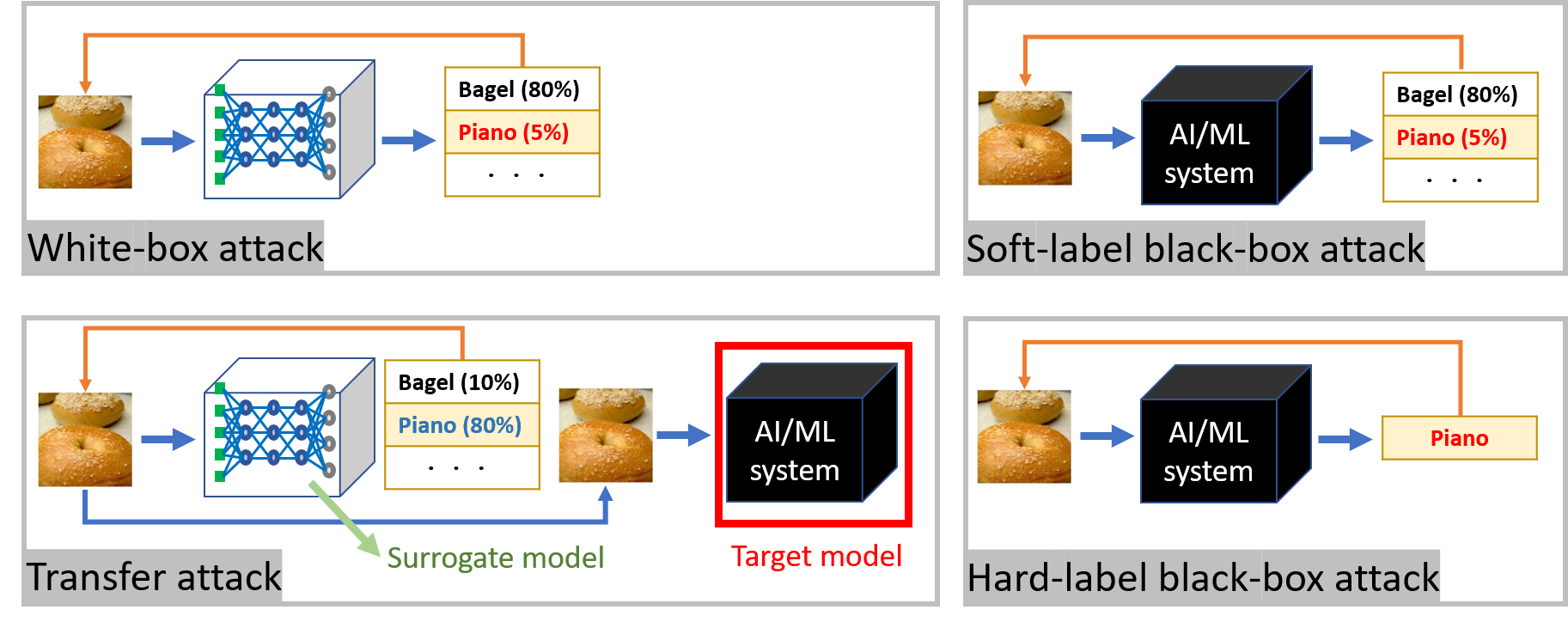}
    \caption{Taxonomy and illustration of evasion attacks. }
    \label{Fig_attack_type}
\end{figure*}

\subsection{Deployment-Phase Attacks}

The objective of deployment-phase attacks is to find a ``similar'' example $\cT(x)$ of $x$ such that the fixed model $f_\theta$ will evade its prediction from the original groundtruth label $y$. The evasion condition can be further separated into two cases: (i) untargeted attack such that  $f_\theta(x)=y$ but $f_\theta(\cT(x)) \neq y$, or (ii) targeted attack such that $f_\theta(x)=y$ but $f_\theta(\cT(x)) = t$, $t \neq y$.
Such $\cT(x)$ is known as an adversarial example\footnote{Sometimes adversarial example may carry a broader meaning of any data sample $\xadv$ that leads to incorrect prediction and therefore dropping the dependence to a reference sample $x$, such as unrestricted adversarial examples \cite{brown2018unrestricted}.} of $x$ \cite{biggio2013evasion,szegedy2013intriguing,goodfellow2014explaining}, and it can be interpreted as out-of-distribution sample or generalization error \cite{stutz2019disentangling}.

\paragraph{Data Similarity.} Depending on data characteristics, specifying a transformation function $\cT(\cdot)$ that preserves data similarity between an original sample $x$ and its transformed sample $\cT(x)$ is a core mission for evasion attacks. The transformation can also be a composite function of semantic-preserving changes \cite{tsai2022towards}.
A common practice to select $\cT(\cdot)$ is through a simple additive perturbation $\delta$ such that $\xadv=x+\delta$, or through domain-specific knowledge such as rotation, object translation, and color changes for image data \cite{hosseini2018semantic,engstrom2019exploring}. For additive perturbation (either on data input or parameter(s) simulating semantic changes), the $\ell_p$ norm ($p \geq 1$) of $\delta$ defined as $\|\delta\|_p  \triangleq \left( \sum_{i=1}^d |\delta_i|^p \right) ^{1/p} $ and the pseudo norm $\ell_0$ are surrogate metrics for measuring similarity distance. Table \ref{tab:similarity} summarizes popular choices of $\ell_p$ norms and their meanings. Take image as an example, $\ell_0$ norm is used to design few-pixel (patch) attacks \cite{su2019one},  $\ell_1$ norm is used to generate sparse and small perturbations \cite{chen2017ead}, $\ell_\infty$ norm is used to confine maximal changes in pixel values \cite{szegedy2013intriguing}, and mixed $\ell_p$ norms can also be used \cite{xu2018structured}.

\paragraph{Evasion Attack Taxonomy.}  Evasion attacks can be categorized based on attackers' knowledge of the target model. Figure \ref{Fig_attack_type} illustrates the taxonomy of different attack types.

\textit{White-box attack} assumes complete knowledge about the target model, including model architecture and model parameters. Consequently, an attacker can exploit the auto differentiation function offered by deep learning packages, such as backpropagation (input gradient) from the model output to the model input, to craft adversarial examples.

\textit{Black-box attack} assumes an attacker can only observe the model prediction of a data input (that is, a query) and does not know any other information.  The target model can be viewed as a black-box function and thus backpropagation for computing input gradient is infeasible without knowing the model details.
In the \textit{soft-label black-box attack} setting, an attacker can observe (parts of) class predictions and their associated confidence scores. In the \textit{hard-label black-box attack} (decision-based) setting, an attacker can only observe the top-1 label prediction, which is the least information required to be returned to remain the utility of the model. In addition to attack success rate, query efficiency is also an important metric for the performance evaluation of black-box attacks.

\textit{Transfer attack} is a branch of black-box attack that uses adversarial examples  generated from a white-box surrogate model to attack the target model. The surrogate model can be either pre-trained \cite{liu2016delving} or distilled from a set of data samples with soft labels given by the target model for training \cite{papernot2016transferability,papernot2017practical}.

\paragraph{Attack formulation.} The process of finding an adversarial perturbation $\delta$ can be formulated as a constrained optimization problem with a specified attack loss $g(\delta|x,y,t,\theta)$ reflecting the attack objective ($t$ is omitted for untargeted attack). The variation in problem formulations and solvers will lead to different attack algorithms. We specify three examples below. Without loss of generality, we use $\ell_p$ norm as the similarity measure (distortion) and untargeted attack as the objective, and assume that all feasible data inputs lie in the scaled space $\cS=[0,1]^d$.

\noindent $\bullet$ Minimal-distortion formulation:
\begin{align}
    \textnormal{Minimize}_{\delta:x+\delta \in \cS}~\|\delta \|_p  \textnormal{~~subject~to~} \yhat_\theta(x+\delta) \neq \yhat_\theta(x) 
\end{align}

\noindent $\bullet$ Penalty-based formulation:
\begin{align}
    \textnormal{Minimize}_{\delta:x+\delta \in \cS}~\|\delta \|_p + \lambda \cdot g(\delta|x,y,\theta)
\end{align}

\noindent $\bullet$ Budget-based (norm bounded)  formulation:
\begin{align}
    \textnormal{Minimize}_{\delta:x+\delta \in \cS}~g(\delta|x,y,\theta)  \textnormal{~~subject~to~} \|\delta\|_p \leq \epsilon
\end{align}
For untargeted attacks, the attacker's loss can be the negative classification loss $g(\delta|x,y,\theta)=-\loss(f_\theta(x+\delta),y)$ or the truncated class margin loss (using either logit or softmax output)  defined as $g(\delta|x,y,\theta)=\max \{  [\logit(x+\delta)]_y - \max_{k\in[K], k \neq y } [\logit(x+\delta)]_k + \kappa  ,0\}$. The margin loss suggests that $g(\delta|x,y,\theta)$ achieves minimal value (i.e. 0) when the top-1 class confidence score excluding the original class $y$ satisfies  $\max_{k\in[K], k \neq y  } [\logit(x+\delta)]_k \geq \logit(x+\delta)]_y + \kappa$, where $\kappa \geq 0$ is a tuning parameter governing their confidence gap. Similarly, for targeted attacks, the attacker's loss can be $g(\delta|x,y,t,\theta)=\loss(f_\theta(x+\delta),t)$ or $g(\delta|x,y,t,\theta)= \max \{  \max_{k\in[K], k \neq t } [\logit(x+\delta)]_k - [\logit(x+\delta)]_t + \kappa  ,0\}$. When implementing black-box attacks, the logit margin loss can be replaced with the observable model output $\log f_\theta$.

The attack formulation can be generalized to the 
\textit{universal perturbation} setting such that it simultaneously evades all model predictions. The universality can be w.r.t. data samples \cite{moosavi2016universal}, model ensembles \cite{tramer2017ensemble}, or various data transformations \cite{athalye2017synthesizing}. \cite{wang2021adversarial} shows that min-max optimization can yield effective universal perturbations.

\paragraph{Selected Attack Algorithms.} We show some white-box and black-box attack algorithms driven by the three aforementioned attack formulations. For the minimal-distortion formulation, the attack constraint $\yhat_\theta(x+\delta) \neq \yhat_\theta(x) $ can be rewritten as $\max_{k\in [K],k \neq y } [f_\theta(x+\delta)]_k \geq [f_\theta(x+\delta)]_y $, which can be used to linearize the local decision boundary around $x$ and allow for efficient projection to the closest linearized decision boundary, leading to white-box attack algorithms such as DeepFool \cite{moosavi2016deepfool} and fast adaptive boundary (FAB) attack \cite{croce2020minimally}.
For the penalty-based formulation, one can use change-of-variable on $\delta$ to convert to an unconstrained optimization problem and then 
use binary search on $\lambda$ to find the smallest $\lambda$ leading to successful attack (i.e., $g=0$), known as Carlini-Wagner (C\&W) white-box attack \cite{carlini2017towards}. For the budget-based formulation, one can apply projected gradient descent (PGD), leading to the white-box PGD attack \cite{madry2017towards}. Attack algorithms using input gradients of the loss function are called gradient-based attacks.

Black-box attack algorithms often adopt either the penalty-based or budget-based formulation. Since the input gradient of the attacker's loss is unavailable to obtain in the black-box setting, one principal approach is to perform gradient estimation using model queries and then use the estimated gradient to replace the true gradient in white-box attack algorithms, leading to the zeroth-order optimization (ZOO) based black-box attacks \cite{chen2017zoo}. 
The choices in gradient estimators \cite{tu2018autozoom,bhagoji2017exploring,liu2018zeroth,liu2018signsgd,ilyas2018black} and ZOO solvers \cite{zhao2019design,zhao2020towards,ilyas2018prior} will give rise to different attack algorithms.
Hard-label black-box attacks can still adopt ZOO principles by spending extra queries to explore local loss landscapes for gradient estimation \cite{cheng2018query,cheng2020signopt,chen2020hopskipjumpattack}, which is more query-efficient than random exploration \cite{brendel2017decision}. We refer the readers to \cite{liu2020primer} for more details on ZOO methods and applications.

\paragraph{Physical adversarial example} is a prediction-evasive physical adversarial object. Examples include stop sign \cite{eykholt2018robust}, eyeglass \cite{sharif2016accessorize}, physical patch \cite{brown2017adversarial},
3D printing \cite{athalye2017synthesizing}, T-shirt \cite{xu2020adversarial}, and facial makeup \cite{lin2021real}.

\section{Defenses and Verification}

Defenses are adversarial threat detection and mitigation strategies, which can be divided into \textit{empirical} and \textit{certified} defenses.  We note that the interplay between attack and defense is essentially a cat-and-mouse game. Many seemingly successful empirical defenses were later weakened by advanced attacks that are defense-aware, which gives a false sense of adversarial robustness due to \textit{information obfuscation} \cite{carlini2017adversarial,athalye2018obfuscated}. Consequently, defenses are expected to be fully evaluated against the best possible \textit{adaptive attacks} that are defense-aware \cite{carlini2019evaluating,tramer2020adaptive}.

While \textit{empirical robustness} refers to the model performance against a set of known attacks, it may fail to serve as a proper robustness indicator against advanced and unseen attacks. To address this issue,  \textit{certified robustness} is used to ensure the model is provably robust given a set of attack conditions (threat models). \textit{Verification} can be viewed as a passive certified defense in the sense that its goal is to quantify a given model's (local) robustness with guarantees.

\subsection{Empirical Defenses}

Empirical defenses are hardening methods applied during the training/deployment phase to improve adversarial robustness without provable guarantees of their effectiveness. 

For training-phase attacks, data filtering and model finetuning are major approaches. For instance, \cite{tran2018spectral} shows removing outliers using learned latent representations and retraining the model can reduce the poison effect. To inspect whether a pre-trained model has a backdoor or not, Neural Cleanse \cite{wang2019neural} reverse-engineers potential trigger patterns for detection. \cite{wang2020practical} proposes data-efficient detectors that require only one sample per class and are made data-free for convolutional neural networks. \cite{Zhao2020Bridging} exploits the mode connectivity in the loss landscape to recover a backdoored model using limited clean data.

For deployment-phase attacks, adversarial input detection schemes that exploit data characteristics such as spatial or temporal correlations are shown to be effective, such as the detection of audio adversarial examples using temporal dependency \cite{yang2018characterizing}. For training adversarially robust models, \textit{adversarial training} that aims to minimize the worst-case loss evaluated by perturbed examples generated during training is so far the strongest empirical defense \cite{madry2017towards}. Specifically, the standard formulation of adversarial training can be expressed as the following min-max optimization over training samples $\{x_i,y_i\}_{i=1}^n$: 
\begin{align}
    \min_{\theta} \sum_{i=1}^n \max_{ \|\delta_i\|_p \leq \epsilon} \loss (f_\theta(x_i+\delta),y_i)
\end{align}
The worst-case loss corresponding to the inner maximization step is often evaluated by gradient-based attacks such as PGD attack \cite{madry2017towards}. Variants of adversarial training methods such as TRADES \cite{zhang2019theoretically} and customized adversarial training (CAT) \cite{cheng2020cat} have been proposed for improved robustness.
\cite{cheng2020self} proposes attack-independent robust training based on self-progression.
On 18 different ImageNet pre-trained models, \cite{su2018robustness} unveils an undesirable trade-off between standard accuracy and adversarial robustness. This trade-off can be improved with unlabeled data \cite{carmon2019unlabeled,stanforth2019labels}. A similar study on vision transformers is presented in \cite{shao2021adversarial}. \cite{paul2022vision} extends the analysis to a variety of robustness aspects beyond adversarial robustness.

\subsection{Certified Defenses}

Certified defenses provide performance guarantees on hardened models.  Adversarial attacks are ineffective if their threat models fall within the provably robust conditions.

For training-phase attacks, \cite{steinhardt2017certified} proposes certified data sanitization against poisoning attacks. \cite{weber2020rab} proposes randomized data training for certified defense against backdoor attacks. For deployment-phase attacks, randomized smoothing is an effective and model-agnostic approach that adds random noises to data input to ``smooth'' the model and perform majority voting on the model predictions. The certified radius (region) in $\ell_p$-norm perturbation ensuring consistent class prediction can be computed by information-theoretical approach \cite{li2018certified}, differential privacy \cite{lecuyer2019certified}, Neyman-Pearson lemma \cite{cohen2019certified}, or higher-order certification \cite{mohapatra2020higher}. The certified defense is also recently extended to robustify black-box victim models by leveraging the technique of denoised randomized smoothing \cite{salman2020denoised,zhang2022how}.

\subsection{Verification}

Verification is often used in certifying local robustness against evasion attacks. Given a neural network $f_\theta$ and a data sample $x$, verification (in its simplest form) aims to maximally certify an $\ell_p$-norm bounded radius $r$ on the perturbation $\delta$ to ensure the model prediction on the perturbed sample $x+\delta$ is consistent as long as $\delta$ is within the certified region. That is, for any $\delta$ such that $\|\delta\|_p \leq r$, $\yhat_\theta(x+\delta)=\yhat_\theta(x)$. The certified radius is a robustness certificate relating to the distance to the closest decision boundary, which is computationally challenging (NP-complete)  for neural networks \cite{katz2017reluplex}.
However, its estimate (hence not a certificate) can be efficiently computed and used as a model-agnostic robustness metric, such as the CLEVER score \cite{weng2018evaluating}. To address the non-linearity induced by layer propagation in neural networks, solving for a certified radius is often cast as a relaxed optimization problem. The methods include convex polytope \cite{wong2018provable}, semidefinite programming \cite{raghunathan2018certified}, dual optimization \cite{dvijotham2018dual}, layer-wise linear bounds \cite{weng2018towards}, and interval bound propagation \cite{gowal2019scalable}. The verification tools are also expanded to support general network architectures \cite{zhang2018efficient,boopathy2018cnn,xu2020automatic} and semantic adversarial examples \cite{mohapatra2020towards}.
The intermediate certified results can be used to train a more certifiable model \cite{wong2018scaling,boopathy2021fast}.
However, scalability to large-sized neural networks remains a major challenge in verification.

\section{Remarks and Discussion}
\label{sec_remark}

Here we make several concluding remarks and discussions.

\paragraph{Novel Applications.} 
The insights from studying adversarial robustness have led to several new use cases. 
Adversarial perturbation and data poisoning are used in generating contrastive explanations \cite{dhurandhar2018explanations},
personal privacy protection \cite{shan2020fawkes}, data/model watermarking and fingerprinting \cite{sablayrolles2020radioactive,aramoon2021don,wangcharacteristi}, data-limited transfer learning \cite{tsai2020transfer,yang2021voice2series}, and visual prompting \cite{bahng2022visual,chen2022understanding,chen2022visual}.
Adversarial examples with proper design are also efficient data augmentation tools to simultaneously improve model generalization  and adversarial robustness \cite{hsu2022adversarial,lei2018discrete}.
Other noteworthy applications include 
image synthesis \cite{santurkar2019image}
generating contrastive explanations \cite{dhurandhar2018explanations},
robust text CAPTCHAs \cite{shao2021robust}, reverse engineering of deception \cite{gong2022reverse},
uncertainty calibration \cite{tang2022neural},  and molecule discovery \cite{hoffman2022optimizing}.

\paragraph{Adversarial Robustness Beyond Classification and Input Perturbation.} The formulations and principles in attacks and defenses for classification can be analogously applied to other machine learning tasks. Examples include sequence-to-sequence translation \cite{cheng2018seq2sick} and image captioning \cite{chen2018attacking}.
Beyond input perturbation, the robustness of model parameter perturbation \cite{tsai2021formalizing} also relates to model quantification \cite{weng2020towards} and energy-efficient inference \cite{stutz2020bit}. 

\paragraph{Instilling Adversarial Robustness into Foundation Models.}
As foundation models \cite{bommasani2021opportunities} adapt task-independent pre-training for general representation learning followed by task-specific fine-tuning for fast adaptation, it is of utmost importance to understand (i) how to incorporate adversarial robustness into foundation model pre-training and (ii) how to maximize adversarial robustness transfer from pre-training to fine-tuning. 
\cite{fan2021does,wang2021fast} show promising results in adversarial robustness preservation and transfer in  meta learning and contrastive learning.
The rapid growth and intensifying demand on foundation models create a unique opportunity to advocate adversarial robustness as a necessary native property in next-generation trustworthy AI tools and call for novel methods for evaluating \textit{representational robustness}, such as in \cite{ko2022synbench}.

\paragraph{Practical Adversarial Robustness at Scale.} From an industrial viewpoint, current solutions to strengthen adversarial robustness may not be ideal because of the unacceptable performance drop on the original task 
and the poor scalability of effective defenses to industry-scale large deep learning models and systems. While there are some efforts for enabling adversarial training at scale, such as \cite{zhang2022distributed}, the notable tradeoff between standard accuracy and robust accuracy may not be a favorable solution for business adoption. An alternative can be rethinking the evaluation methodology of adversarial robustness. For example, instead of aiming to mitigate the robustness-accuracy tradeoff, we can compare the unilateral robustness gain under the constraint of making minimal (or even zero) harm to the original model utility (e.g. test accuracy). Moreover, an ideal defense should be lightweight and deployable in a plug-and-play manner for any given model, instead of demanding to train a model from scratch for improved robustness. 


\bibliography{adversarial_learning_short_20210227}

\begin{thebibliography}{124}
\providecommand{\natexlab}[1]{#1}

\bibitem[{Aramoon, Chen, and Qu(2021)}]{aramoon2021don}
Aramoon, O.; Chen, P.-Y.; and Qu, G. 2021.
\newblock Don't Forget to Sign the Gradients!
\newblock \emph{MLSyS}, 3.

\bibitem[{Athalye, Carlini, and Wagner(2018)}]{athalye2018obfuscated}
Athalye, A.; Carlini, N.; and Wagner, D. 2018.
\newblock Obfuscated gradients give a false sense of security: Circumventing
  defenses to adversarial examples.
\newblock \emph{ICML}.

\bibitem[{Athalye and Sutskever(2018)}]{athalye2017synthesizing}
Athalye, A.; and Sutskever, I. 2018.
\newblock Synthesizing robust adversarial examples.
\newblock \emph{ICML}.

\bibitem[{Bagdasaryan et~al.(2020)Bagdasaryan, Veit, Hua, Estrin, and
  Shmatikov}]{bagdasaryan2020backdoor}
Bagdasaryan, E.; Veit, A.; Hua, Y.; Estrin, D.; and Shmatikov, V. 2020.
\newblock How to backdoor federated learning.
\newblock \emph{AISTATS}.

\bibitem[{Bahng et~al.(2022)Bahng, Jahanian, Sankaranarayanan, and
  Isola}]{bahng2022visual}
Bahng, H.; Jahanian, A.; Sankaranarayanan, S.; and Isola, P. 2022.
\newblock Visual Prompting: Modifying Pixel Space to Adapt Pre-trained Models.
\newblock \emph{arXiv preprint arXiv:2203.17274}.

\bibitem[{Bhagoji et~al.(2019)Bhagoji, Chakraborty, Mittal, and
  Calo}]{bhagoji2019analyzing}
Bhagoji, A.~N.; Chakraborty, S.; Mittal, P.; and Calo, S. 2019.
\newblock Analyzing federated learning through an adversarial lens.
\newblock \emph{ICML}, 634--643.

\bibitem[{Biggio et~al.(2013)Biggio, Corona, Maiorca, Nelson, {\v{S}}rndi{\'c},
  Laskov, Giacinto, and Roli}]{biggio2013evasion}
Biggio, B.; Corona, I.; Maiorca, D.; Nelson, B.; {\v{S}}rndi{\'c}, N.; Laskov,
  P.; Giacinto, G.; and Roli, F. 2013.
\newblock Evasion attacks against machine learning at test time.
\newblock \emph{ECML PKDD}.

\bibitem[{Biggio and Roli(2018)}]{biggio2018wild}
Biggio, B.; and Roli, F. 2018.
\newblock Wild patterns: Ten years after the rise of adversarial machine
  learning.
\newblock \emph{Pattern Recognition}, 84: 317--331.

\bibitem[{Bommasani et~al.(2021)Bommasani, Hudson, Adeli, Altman, Arora, von
  Arx, Bernstein, Bohg, Bosselut, Brunskill
  et~al.}]{bommasani2021opportunities}
Bommasani, R.; Hudson, D.~A.; Adeli, E.; Altman, R.; Arora, S.; von Arx, S.;
  Bernstein, M.~S.; Bohg, J.; Bosselut, A.; Brunskill, E.; et~al. 2021.
\newblock On the opportunities and risks of foundation models.
\newblock \emph{arXiv preprint arXiv:2108.07258}.

\bibitem[{Boopathy et~al.(2019)Boopathy, Weng, Chen, Liu, and
  Daniel}]{boopathy2018cnn}
Boopathy, A.; Weng, T.-W.; Chen, P.-Y.; Liu, S.; and Daniel, L. 2019.
\newblock {CNN}-cert: An efficient framework for certifying robustness of
  convolutional neural networks.
\newblock \emph{AAAI}.

\bibitem[{Boopathy et~al.(2021)Boopathy, Weng, Liu, Chen, Zhang, and
  Daniel}]{boopathy2021fast}
Boopathy, A.; Weng, T.-W.; Liu, S.; Chen, P.-Y.; Zhang, G.; and Daniel, L.
  2021.
\newblock Fast Training of Provably Robust Neural Networks by SingleProp.
\newblock \emph{AAAI}.

\bibitem[{Brendel, Rauber, and Bethge(2018)}]{brendel2017decision}
Brendel, W.; Rauber, J.; and Bethge, M. 2018.
\newblock Decision-Based Adversarial Attacks: Reliable Attacks Against
  Black-Box Machine Learning Models.
\newblock \emph{ICLR}.

\bibitem[{Brown et~al.(2018)Brown, Carlini, Zhang, Olsson, Christiano, and
  Goodfellow}]{brown2018unrestricted}
Brown, T.~B.; Carlini, N.; Zhang, C.; Olsson, C.; Christiano, P.; and
  Goodfellow, I. 2018.
\newblock Unrestricted adversarial examples.
\newblock \emph{arXiv preprint arXiv:1809.08352}.

\bibitem[{Brown et~al.(2017)Brown, Man{\'e}, Roy, Abadi, and
  Gilmer}]{brown2017adversarial}
Brown, T.~B.; Man{\'e}, D.; Roy, A.; Abadi, M.; and Gilmer, J. 2017.
\newblock Adversarial patch.
\newblock \emph{arXiv preprint arXiv:1712.09665}.

\bibitem[{Carlini et~al.(2019{\natexlab{a}})Carlini, Athalye, Papernot,
  Brendel, Rauber, Tsipras, Goodfellow, Madry, and
  Kurakin}]{carlini2019evaluating}
Carlini, N.; Athalye, A.; Papernot, N.; Brendel, W.; Rauber, J.; Tsipras, D.;
  Goodfellow, I.; Madry, A.; and Kurakin, A. 2019{\natexlab{a}}.
\newblock On evaluating adversarial robustness.
\newblock \emph{arXiv preprint arXiv:1902.06705}.

\bibitem[{Carlini et~al.(2019{\natexlab{b}})Carlini, Liu, Erlingsson, Kos, and
  Song}]{carlini2019secret}
Carlini, N.; Liu, C.; Erlingsson, {\'U}.; Kos, J.; and Song, D.
  2019{\natexlab{b}}.
\newblock The secret sharer: Evaluating and testing unintended memorization in
  neural networks.
\newblock \emph{USENIX Security}, 267--284.

\bibitem[{Carlini and Wagner(2017{\natexlab{a}})}]{carlini2017adversarial}
Carlini, N.; and Wagner, D. 2017{\natexlab{a}}.
\newblock Adversarial examples are not easily detected: Bypassing ten detection
  methods.
\newblock \emph{ACM Workshop on Artificial Intelligence and Security}, 3--14.

\bibitem[{Carlini and Wagner(2017{\natexlab{b}})}]{carlini2017towards}
Carlini, N.; and Wagner, D. 2017{\natexlab{b}}.
\newblock Towards evaluating the robustness of neural networks.
\newblock \emph{IEEE S\&P}, 39--57.

\bibitem[{Carmon et~al.(2019)Carmon, Raghunathan, Schmidt, Liang, and
  Duchi}]{carmon2019unlabeled}
Carmon, Y.; Raghunathan, A.; Schmidt, L.; Liang, P.; and Duchi, J.~C. 2019.
\newblock Unlabeled data improves adversarial robustness.
\newblock \emph{NeurIPS}.

\bibitem[{Chen et~al.(2022{\natexlab{a}})Chen, Lorenz, Yao, Chen, and
  Liu}]{chen2022visual}
Chen, A.; Lorenz, P.; Yao, Y.; Chen, P.-Y.; and Liu, S. 2022{\natexlab{a}}.
\newblock Visual Prompting for Adversarial Robustness.
\newblock \emph{arXiv preprint arXiv:2210.06284}.

\bibitem[{Chen et~al.(2022{\natexlab{b}})Chen, Yao, Chen, Zhang, and
  Liu}]{chen2022understanding}
Chen, A.; Yao, Y.; Chen, P.-Y.; Zhang, Y.; and Liu, S. 2022{\natexlab{b}}.
\newblock Understanding and Improving Visual Prompting: A Label-Mapping
  Perspective.
\newblock \emph{arXiv preprint arXiv:2211.11635}.

\bibitem[{Chen et~al.(2018{\natexlab{a}})Chen, Zhang, Chen, Yi, and
  Hsieh}]{chen2018attacking}
Chen, H.; Zhang, H.; Chen, P.-Y.; Yi, J.; and Hsieh, C.-J. 2018{\natexlab{a}}.
\newblock Attacking visual language grounding with adversarial examples: A case
  study on neural image captioning.
\newblock \emph{ACL}, 1: 2587--2597.

\bibitem[{Chen, Jordan, and Wainwright(2020)}]{chen2020hopskipjumpattack}
Chen, J.; Jordan, M.~I.; and Wainwright, M.~J. 2020.
\newblock Hopskipjumpattack: A query-efficient decision-based attack.
\newblock \emph{IEEE S\&P}.

\bibitem[{Chen et~al.(2018{\natexlab{b}})Chen, Sharma, Zhang, Yi, and
  Hsieh}]{chen2017ead}
Chen, P.-Y.; Sharma, Y.; Zhang, H.; Yi, J.; and Hsieh, C.-J.
  2018{\natexlab{b}}.
\newblock {EAD}: elastic-net attacks to deep neural networks via adversarial
  examples.
\newblock \emph{AAAI}, 10--17.

\bibitem[{Chen et~al.(2017)Chen, Zhang, Sharma, Yi, and Hsieh}]{chen2017zoo}
Chen, P.-Y.; Zhang, H.; Sharma, Y.; Yi, J.; and Hsieh, C.-J. 2017.
\newblock {ZOO}: Zeroth Order Optimization Based Black-box Attacks to Deep
  Neural Networks Without Training Substitute Models.
\newblock \emph{ACM Workshop on Artificial Intelligence and Security}, 15--26.

\bibitem[{Cheng et~al.(2021)Cheng, Chen, Liu, Chang, Hsieh, and
  Das}]{cheng2020self}
Cheng, M.; Chen, P.-Y.; Liu, S.; Chang, S.; Hsieh, C.-J.; and Das, P. 2021.
\newblock Self-Progressing Robust Training.
\newblock \emph{AAAI}.

\bibitem[{Cheng et~al.(2019)Cheng, Le, Chen, Yi, Zhang, and
  Hsieh}]{cheng2018query}
Cheng, M.; Le, T.; Chen, P.-Y.; Yi, J.; Zhang, H.; and Hsieh, C.-J. 2019.
\newblock Query-efficient hard-label black-box attack: An optimization-based
  approach.
\newblock \emph{ICLR}.

\bibitem[{Cheng et~al.(2020{\natexlab{a}})Cheng, Lei, Chen, Dhillon, and
  Hsieh}]{cheng2020cat}
Cheng, M.; Lei, Q.; Chen, P.-Y.; Dhillon, I.; and Hsieh, C.-J.
  2020{\natexlab{a}}.
\newblock Cat: Customized adversarial training for improved robustness.
\newblock \emph{arXiv preprint arXiv:2002.06789}.

\bibitem[{Cheng et~al.(2020{\natexlab{b}})Cheng, Singh, Chen, Chen, Liu, and
  Hsieh}]{cheng2020signopt}
Cheng, M.; Singh, S.; Chen, P.~H.; Chen, P.-Y.; Liu, S.; and Hsieh, C.-J.
  2020{\natexlab{b}}.
\newblock Sign-{OPT}: A Query-Efficient Hard-label Adversarial Attack.
\newblock \emph{ICLR}.

\bibitem[{Cheng et~al.(2020{\natexlab{c}})Cheng, Yi, Zhang, Chen, and
  Hsieh}]{cheng2018seq2sick}
Cheng, M.; Yi, J.; Zhang, H.; Chen, P.-Y.; and Hsieh, C.-J. 2020{\natexlab{c}}.
\newblock Seq2Sick: Evaluating the Robustness of Sequence-to-Sequence Models
  with Adversarial Examples.
\newblock \emph{AAAI}.

\bibitem[{Cohen, Rosenfeld, and Kolter(2019)}]{cohen2019certified}
Cohen, J.~M.; Rosenfeld, E.; and Kolter, J.~Z. 2019.
\newblock Certified adversarial robustness via randomized smoothing.
\newblock \emph{ICML}.

\bibitem[{Croce and Hein(2020)}]{croce2020minimally}
Croce, F.; and Hein, M. 2020.
\newblock Minimally distorted adversarial examples with a fast adaptive
  boundary attack.
\newblock \emph{ICML}, 2196--2205.

\bibitem[{Dhurandhar et~al.(2018)Dhurandhar, Chen, Luss, Tu, Ting, Shanmugam,
  and Das}]{dhurandhar2018explanations}
Dhurandhar, A.; Chen, P.-Y.; Luss, R.; Tu, C.-C.; Ting, P.; Shanmugam, K.; and
  Das, P. 2018.
\newblock Explanations based on the missing: Towards contrastive explanations
  with pertinent negatives.
\newblock \emph{NeurIPS}.

\bibitem[{Dvijotham et~al.(2018)Dvijotham, Stanforth, Gowal, Mann, and
  Kohli}]{dvijotham2018dual}
Dvijotham, K.; Stanforth, R.; Gowal, S.; Mann, T.~A.; and Kohli, P. 2018.
\newblock A Dual Approach to Scalable Verification of Deep Networks.
\newblock \emph{UAI}, 1(2): 3.

\bibitem[{Engstrom et~al.(2019)Engstrom, Tran, Tsipras, Schmidt, and
  Madry}]{engstrom2019exploring}
Engstrom, L.; Tran, B.; Tsipras, D.; Schmidt, L.; and Madry, A. 2019.
\newblock Exploring the landscape of spatial robustness.
\newblock \emph{ICML}, 1802--1811.

\bibitem[{Eykholt et~al.(2018)Eykholt, Evtimov, Fernandes, Li, Rahmati, Xiao,
  Prakash, Kohno, and Song}]{eykholt2018robust}
Eykholt, K.; Evtimov, I.; Fernandes, E.; Li, B.; Rahmati, A.; Xiao, C.;
  Prakash, A.; Kohno, T.; and Song, D. 2018.
\newblock Robust physical-world attacks on deep learning visual classification.
\newblock \emph{CVPR}, 1625--1634.

\bibitem[{Fan et~al.(2021)Fan, Liu, Chen, Zhang, and Gan}]{fan2021does}
Fan, L.; Liu, S.; Chen, P.-Y.; Zhang, G.; and Gan, C. 2021.
\newblock When Does Contrastive Learning Preserve Adversarial Robustness from
  Pretraining to Finetuning?
\newblock \emph{NeurIPS}, 34.

\bibitem[{Geiping et~al.(2021)Geiping, Fowl, Huang, Czaja, Taylor, Moeller, and
  Goldstein}]{geiping2020witches}
Geiping, J.; Fowl, L.; Huang, W.~R.; Czaja, W.; Taylor, G.; Moeller, M.; and
  Goldstein, T. 2021.
\newblock Witches' Brew: Industrial Scale Data Poisoning via Gradient Matching.
\newblock \emph{ICLR}.

\bibitem[{Goldblum et~al.(2020)Goldblum, Tsipras, Xie, Chen, Schwarzschild,
  Song, Madry, Li, and Goldstein}]{goldblum2020data}
Goldblum, M.; Tsipras, D.; Xie, C.; Chen, X.; Schwarzschild, A.; Song, D.;
  Madry, A.; Li, B.; and Goldstein, T. 2020.
\newblock Data Security for Machine Learning: Data Poisoning, Backdoor Attacks,
  and Defenses.
\newblock \emph{arXiv preprint arXiv:2012.10544}.

\bibitem[{Gong et~al.(2022)Gong, Yao, Li, Zhang, Liu, Lin, and
  Liu}]{gong2022reverse}
Gong, Y.; Yao, Y.; Li, Y.; Zhang, Y.; Liu, X.; Lin, X.; and Liu, S. 2022.
\newblock Reverse Engineering of Imperceptible Adversarial Image Perturbations.
\newblock \emph{arXiv preprint arXiv:2203.14145}.

\bibitem[{Goodfellow, Shlens, and Szegedy(2015)}]{goodfellow2014explaining}
Goodfellow, I.~J.; Shlens, J.; and Szegedy, C. 2015.
\newblock Explaining and harnessing adversarial examples.
\newblock \emph{ICLR}.

\bibitem[{Gowal et~al.(2019)Gowal, Dvijotham, Stanforth, Bunel, Qin, Uesato,
  Arandjelovic, Mann, and Kohli}]{gowal2019scalable}
Gowal, S.; Dvijotham, K.~D.; Stanforth, R.; Bunel, R.; Qin, C.; Uesato, J.;
  Arandjelovic, R.; Mann, T.; and Kohli, P. 2019.
\newblock Scalable verified training for provably robust image classification.
\newblock \emph{ICCV}, 4842--4851.

\bibitem[{{Gu} et~al.(2019){Gu}, {Liu}, {Dolan-Gavitt}, and
  {Garg}}]{BadNet_Access}
{Gu}, T.; {Liu}, K.; {Dolan-Gavitt}, B.; and {Garg}, S. 2019.
\newblock {BadNets}: Evaluating Backdooring Attacks on Deep Neural Networks.
\newblock \emph{IEEE Access}, 7: 47230--47244.

\bibitem[{Hoffman et~al.(2022)Hoffman, Chenthamarakshan, Wadhawan, Chen, and
  Das}]{hoffman2022optimizing}
Hoffman, S.~C.; Chenthamarakshan, V.; Wadhawan, K.; Chen, P.-Y.; and Das, P.
  2022.
\newblock Optimizing molecules using efficient queries from property
  evaluations.
\newblock \emph{Nature Machine Intelligence}, 4(1): 21--31.

\bibitem[{Hosseini and Poovendran(2018)}]{hosseini2018semantic}
Hosseini, H.; and Poovendran, R. 2018.
\newblock Semantic adversarial examples.
\newblock \emph{CVPR Workshops}, 1614--1619.

\bibitem[{Hsiung et~al.(2022)Hsiung, Tsai, Chen, and Ho}]{tsai2022towards}
Hsiung, L.; Tsai, Y.-Y.; Chen, P.-Y.; and Ho, T.-Y. 2022.
\newblock Towards Compositional Adversarial Robustness: Generalizing
  Adversarial Training to Composite Semantic Perturbations.
\newblock \emph{arXiv preprint arXiv:2202.04235}.

\bibitem[{Hsu et~al.(2022)Hsu, Chen, Lu, Liu, and Yu}]{hsu2022adversarial}
Hsu, C.-Y.; Chen, P.-Y.; Lu, S.; Liu, S.; and Yu, C.-M. 2022.
\newblock Adversarial Examples can be Effective Data Augmentation for
  Unsupervised Machine Learning.
\newblock In \emph{AAAI}.

\bibitem[{Ilyas et~al.(2018)Ilyas, Engstrom, Athalye, and Lin}]{ilyas2018black}
Ilyas, A.; Engstrom, L.; Athalye, A.; and Lin, J. 2018.
\newblock Black-box Adversarial Attacks with Limited Queries and Information.
\newblock \emph{ICML}.

\bibitem[{Ilyas, Engstrom, and Madry(2019)}]{ilyas2018prior}
Ilyas, A.; Engstrom, L.; and Madry, A. 2019.
\newblock Prior convictions: Black-box adversarial attacks with bandits and
  priors.
\newblock \emph{ICLR}.

\bibitem[{Jagielski et~al.(2018)Jagielski, Oprea, Biggio, Liu, Nita-Rotaru, and
  Li}]{jagielski2018manipulating}
Jagielski, M.; Oprea, A.; Biggio, B.; Liu, C.; Nita-Rotaru, C.; and Li, B.
  2018.
\newblock Manipulating machine learning: Poisoning attacks and countermeasures
  for regression learning.
\newblock \emph{IEEE S\&P}, 19--35.

\bibitem[{Katz et~al.(2017)Katz, Barrett, Dill, Julian, and
  Kochenderfer}]{katz2017reluplex}
Katz, G.; Barrett, C.; Dill, D.~L.; Julian, K.; and Kochenderfer, M.~J. 2017.
\newblock Reluplex: An efficient SMT solver for verifying deep neural networks.
\newblock \emph{International Conference on Computer Aided Verification},
  97--117.

\bibitem[{Ko et~al.(2022)Ko, Chen, Mohapatra, Das, and Daniel}]{ko2022synbench}
Ko, C.-Y.; Chen, P.-Y.; Mohapatra, J.; Das, P.; and Daniel, L. 2022.
\newblock SynBench: Task-Agnostic Benchmarking of Pretrained Representations
  using Synthetic Data.
\newblock \emph{arXiv preprint arXiv:2210.02989}.

\bibitem[{Kumar et~al.(2020)Kumar, Nystr{\"o}m, Lambert, Marshall, Goertzel,
  Comissoneru, Swann, and Xia}]{kumar2020adversarial}
Kumar, R. S.~S.; Nystr{\"o}m, M.; Lambert, J.; Marshall, A.; Goertzel, M.;
  Comissoneru, A.; Swann, M.; and Xia, S. 2020.
\newblock Adversarial machine learning-industry perspectives.
\newblock \emph{IEEE S\&P Workshops}, 69--75.

\bibitem[{LeCun, Bengio, and Hinton(2015)}]{lecun2015deep}
LeCun, Y.; Bengio, Y.; and Hinton, G. 2015.
\newblock Deep learning.
\newblock \emph{Nature}.

\bibitem[{Lecuyer et~al.(2019)Lecuyer, Atlidakis, Geambasu, Hsu, and
  Jana}]{lecuyer2019certified}
Lecuyer, M.; Atlidakis, V.; Geambasu, R.; Hsu, D.; and Jana, S. 2019.
\newblock Certified robustness to adversarial examples with differential
  privacy.
\newblock \emph{IEEE S\&P}, 656--672.

\bibitem[{Lei et~al.(2019)Lei, Wu, Chen, Dimakis, Dhillon, and
  Witbrock}]{lei2018discrete}
Lei, Q.; Wu, L.; Chen, P.-Y.; Dimakis, A.~G.; Dhillon, I.~S.; and Witbrock, M.
  2019.
\newblock Discrete Adversarial Attacks and Submodular Optimization with
  Applications to Text Classification.
\newblock \emph{SysML}.

\bibitem[{Li et~al.(2019)Li, Chen, Wang, and Carin}]{li2018certified}
Li, B.; Chen, C.; Wang, W.; and Carin, L. 2019.
\newblock Certified adversarial robustness with additive noise.
\newblock \emph{NeurIPS}.

\bibitem[{Lin et~al.(2022)Lin, Hsu, Chen, and Yu}]{lin2021real}
Lin, C.-S.; Hsu, C.-Y.; Chen, P.-Y.; and Yu, C.-M. 2022.
\newblock Real-World Adversarial Examples Via Makeup.
\newblock In \emph{IEEE International Conference on Acoustics, Speech and
  Signal Processing (ICASSP)}, 2854--2858.

\bibitem[{Liu et~al.(2019)Liu, Chen, Chen, and Hong}]{liu2018signsgd}
Liu, S.; Chen, P.-Y.; Chen, X.; and Hong, M. 2019.
\newblock {signSGD} via Zeroth-Order Oracle.
\newblock \emph{ICLR}.

\bibitem[{Liu et~al.(2020{\natexlab{a}})Liu, Chen, Kailkhura, Zhang, Hero, and
  Varshney}]{liu2020primer}
Liu, S.; Chen, P.-Y.; Kailkhura, B.; Zhang, G.; Hero, A.; and Varshney, P.~K.
  2020{\natexlab{a}}.
\newblock A Primer on Zeroth-Order Optimization in Signal Processing and
  Machine Learning.
\newblock \emph{IEEE Signal Processing Magazine}.

\bibitem[{Liu et~al.(2018)Liu, Kailkhura, Chen, Ting, Chang, and
  Amini}]{liu2018zeroth}
Liu, S.; Kailkhura, B.; Chen, P.-Y.; Ting, P.; Chang, S.; and Amini, L. 2018.
\newblock Zeroth-order stochastic variance reduction for nonconvex
  optimization.
\newblock \emph{NeurIPS}, 3731--3741.

\bibitem[{Liu et~al.(2020{\natexlab{b}})Liu, Lu, Chen, Feng, Xu, Al-Dujaili,
  Hong, and O’Reilly}]{liu2020min}
Liu, S.; Lu, S.; Chen, X.; Feng, Y.; Xu, K.; Al-Dujaili, A.; Hong, M.; and
  O’Reilly, U.-M. 2020{\natexlab{b}}.
\newblock Min-max optimization without gradients: Convergence and applications
  to black-box evasion and poisoning attacks.
\newblock \emph{ICML}, 6282--6293.

\bibitem[{Liu et~al.(2017)Liu, Chen, Liu, and Song}]{liu2016delving}
Liu, Y.; Chen, X.; Liu, C.; and Song, D. 2017.
\newblock Delving into transferable adversarial examples and black-box attacks.
\newblock \emph{ICLR}.

\bibitem[{Madry et~al.(2018)Madry, Makelov, Schmidt, Tsipras, and
  Vladu}]{madry2017towards}
Madry, A.; Makelov, A.; Schmidt, L.; Tsipras, D.; and Vladu, A. 2018.
\newblock Towards Deep Learning Models Resistant to Adversarial Attacks.
\newblock \emph{ICLR}.

\bibitem[{Mohapatra et~al.(2020{\natexlab{a}})Mohapatra, Ko, Weng, Chen, Liu,
  and Daniel}]{mohapatra2020higher}
Mohapatra, J.; Ko, C.-Y.; Weng, T.-W.; Chen, P.-Y.; Liu, S.; and Daniel, L.
  2020{\natexlab{a}}.
\newblock Higher-Order Certification for Randomized Smoothing.
\newblock \emph{NeurIPS}.

\bibitem[{Mohapatra et~al.(2020{\natexlab{b}})Mohapatra, Weng, Chen, Liu, and
  Daniel}]{mohapatra2020towards}
Mohapatra, J.; Weng, T.-W.; Chen, P.-Y.; Liu, S.; and Daniel, L.
  2020{\natexlab{b}}.
\newblock Towards verifying robustness of neural networks against a family of
  semantic perturbations.
\newblock \emph{CVPR}, 244--252.

\bibitem[{Moosavi-Dezfooli et~al.(2017)Moosavi-Dezfooli, Fawzi, Fawzi, and
  Frossard}]{moosavi2016universal}
Moosavi-Dezfooli, S.-M.; Fawzi, A.; Fawzi, O.; and Frossard, P. 2017.
\newblock Universal Adversarial Perturbations.
\newblock \emph{CVPR}, 86--94.

\bibitem[{Moosavi-Dezfooli, Fawzi, and Frossard(2016)}]{moosavi2016deepfool}
Moosavi-Dezfooli, S.-M.; Fawzi, A.; and Frossard, P. 2016.
\newblock Deepfool: a simple and accurate method to fool deep neural networks.
\newblock \emph{CVPR}, 2574--2582.

\bibitem[{Nguyen and Tran(2020)}]{nguyen2020input}
Nguyen, A.; and Tran, A. 2020.
\newblock Input-aware dynamic backdoor attack.
\newblock \emph{NeurIPS}.

\bibitem[{Nitin~Bhagoji et~al.(2018)Nitin~Bhagoji, He, Li, and
  Song}]{bhagoji2017exploring}
Nitin~Bhagoji, A.; He, W.; Li, B.; and Song, D. 2018.
\newblock Practical Black-box Attacks on Deep Neural Networks using Efficient
  Query Mechanisms.
\newblock \emph{ECCV}, 154--169.

\bibitem[{Papernot, McDaniel, and
  Goodfellow(2016)}]{papernot2016transferability}
Papernot, N.; McDaniel, P.; and Goodfellow, I. 2016.
\newblock Transferability in machine learning: from phenomena to black-box
  attacks using adversarial samples.
\newblock \emph{arXiv preprint arXiv:1605.07277}.

\bibitem[{Papernot et~al.(2017)Papernot, McDaniel, Goodfellow, Jha, Celik, and
  Swami}]{papernot2017practical}
Papernot, N.; McDaniel, P.; Goodfellow, I.; Jha, S.; Celik, Z.~B.; and Swami,
  A. 2017.
\newblock Practical black-box attacks against machine learning.
\newblock \emph{ACM Asia Conference on Computer and Communications Security},
  506--519.

\bibitem[{Paul and Chen(2022)}]{paul2022vision}
Paul, S.; and Chen, P.-Y. 2022.
\newblock Vision transformers are robust learners.
\newblock In \emph{Proceedings of the AAAI Conference on Artificial
  Intelligence}, volume~36, 2071--2081.

\bibitem[{Raghunathan, Steinhardt, and Liang(2018)}]{raghunathan2018certified}
Raghunathan, A.; Steinhardt, J.; and Liang, P. 2018.
\newblock Certified a against adversarial examples.
\newblock \emph{ICLR}.

\bibitem[{Sablayrolles et~al.(2020)Sablayrolles, Douze, Schmid, and
  J{\'e}gou}]{sablayrolles2020radioactive}
Sablayrolles, A.; Douze, M.; Schmid, C.; and J{\'e}gou, H. 2020.
\newblock Radioactive data: tracing through training.
\newblock \emph{ICML}.

\bibitem[{Salman et~al.(2020)Salman, Sun, Yang, Kapoor, and
  Kolter}]{salman2020denoised}
Salman, H.; Sun, M.; Yang, G.; Kapoor, A.; and Kolter, J.~Z. 2020.
\newblock Denoised smoothing: A provable defense for pretrained classifiers.
\newblock \emph{Advances in Neural Information Processing Systems}, 33:
  21945--21957.

\bibitem[{Santurkar et~al.(2019)Santurkar, Tsipras, Tran, Ilyas, Engstrom, and
  Madry}]{santurkar2019image}
Santurkar, S.; Tsipras, D.; Tran, B.; Ilyas, A.; Engstrom, L.; and Madry, A.
  2019.
\newblock Image synthesis with a single (robust) classifier.
\newblock \emph{arXiv preprint arXiv:1906.09453}.

\bibitem[{Shafahi et~al.(2018)Shafahi, Huang, Najibi, Suciu, Studer, Dumitras,
  and Goldstein}]{shafahi2018poison}
Shafahi, A.; Huang, W.~R.; Najibi, M.; Suciu, O.; Studer, C.; Dumitras, T.; and
  Goldstein, T. 2018.
\newblock Poison frogs! targeted clean-label poisoning attacks on neural
  networks.
\newblock \emph{NeurIPS}, 6103--6113.

\bibitem[{Shan et~al.(2020)Shan, Wenger, Zhang, Li, Zheng, and
  Zhao}]{shan2020fawkes}
Shan, S.; Wenger, E.; Zhang, J.; Li, H.; Zheng, H.; and Zhao, B.~Y. 2020.
\newblock Fawkes: Protecting privacy against unauthorized deep learning models.
\newblock \emph{USENIX Security}.

\bibitem[{Shao et~al.(2021)Shao, Shi, Yi, Chen, and Hsieh}]{shao2021robust}
Shao, R.; Shi, Z.; Yi, J.; Chen, P.-Y.; and Hsieh, C.-J. 2021.
\newblock Robust Text CAPTCHAs Using Adversarial Examples.
\newblock \emph{arXiv preprint arXiv:2101.02483}.

\bibitem[{Shao et~al.(2022)Shao, Shi, Yi, Chen, and
  Hsieh}]{shao2021adversarial}
Shao, R.; Shi, Z.; Yi, J.; Chen, P.-Y.; and Hsieh, C.-J. 2022.
\newblock On the Adversarial Robustness of Vision Transformers.
\newblock \emph{Transactions on Machine Learning Research}.

\bibitem[{Sharif et~al.(2016)Sharif, Bhagavatula, Bauer, and
  Reiter}]{sharif2016accessorize}
Sharif, M.; Bhagavatula, S.; Bauer, L.; and Reiter, M.~K. 2016.
\newblock Accessorize to a crime: Real and stealthy attacks on state-of-the-art
  face recognition.
\newblock \emph{ACM CCS}, 1528--1540.

\bibitem[{Stanforth et~al.(2019)Stanforth, Fawzi, Kohli
  et~al.}]{stanforth2019labels}
Stanforth, R.; Fawzi, A.; Kohli, P.; et~al. 2019.
\newblock Are Labels Required for Improving Adversarial Robustness?
\newblock \emph{NeurIPS}.

\bibitem[{Steinhardt, Koh, and Liang(2017)}]{steinhardt2017certified}
Steinhardt, J.; Koh, P.~W.; and Liang, P. 2017.
\newblock Certified defenses for data poisoning attacks.
\newblock \emph{NeurIPS}.

\bibitem[{Stutz et~al.(2020)Stutz, Chandramoorthy, Hein, and
  Schiele}]{stutz2020bit}
Stutz, D.; Chandramoorthy, N.; Hein, M.; and Schiele, B. 2020.
\newblock Bit Error Robustness for Energy-Efficient DNN Accelerators.
\newblock \emph{arXiv preprint arXiv:2006.13977}.

\bibitem[{Stutz, Hein, and Schiele(2019)}]{stutz2019disentangling}
Stutz, D.; Hein, M.; and Schiele, B. 2019.
\newblock Disentangling adversarial robustness and generalization.
\newblock \emph{CVPR}, 6976--6987.

\bibitem[{Su et~al.(2018)Su, Zhang, Chen, Yi, Chen, and Gao}]{su2018robustness}
Su, D.; Zhang, H.; Chen, H.; Yi, J.; Chen, P.-Y.; and Gao, Y. 2018.
\newblock Is robustness the cost of accuracy? A comprehensive study on the
  robustness of 18 deep image classification models.
\newblock \emph{ECCV}.

\bibitem[{Su, Vargas, and Sakurai(2019)}]{su2019one}
Su, J.; Vargas, D.~V.; and Sakurai, K. 2019.
\newblock One pixel attack for fooling deep neural networks.
\newblock \emph{IEEE Transactions on Evolutionary Computation}, 23(5):
  828--841.

\bibitem[{Szegedy et~al.(2014)Szegedy, Zaremba, Sutskever, Bruna, Erhan,
  Goodfellow, and Fergus}]{szegedy2013intriguing}
Szegedy, C.; Zaremba, W.; Sutskever, I.; Bruna, J.; Erhan, D.; Goodfellow, I.;
  and Fergus, R. 2014.
\newblock Intriguing properties of neural networks.
\newblock \emph{ICLR}.

\bibitem[{Tang, Chen, and Ho(2022)}]{tang2022neural}
Tang, Y.-C.; Chen, P.-Y.; and Ho, T.-Y. 2022.
\newblock Neural Clamping: Joint Input Perturbation and Temperature Scaling for
  Neural Network Calibration.
\newblock \emph{arXiv preprint arXiv:2209.11604}.

\bibitem[{Tramer et~al.(2020)Tramer, Carlini, Brendel, and
  Madry}]{tramer2020adaptive}
Tramer, F.; Carlini, N.; Brendel, W.; and Madry, A. 2020.
\newblock On adaptive attacks to adversarial example defenses.
\newblock \emph{NeurIPS}.

\bibitem[{Tram{\`e}r et~al.(2018)Tram{\`e}r, Kurakin, Papernot, Boneh, and
  McDaniel}]{tramer2017ensemble}
Tram{\`e}r, F.; Kurakin, A.; Papernot, N.; Boneh, D.; and McDaniel, P. 2018.
\newblock Ensemble Adversarial Training: Attacks and Defenses.
\newblock \emph{ICLR}.

\bibitem[{Tran, Li, and Madry(2018)}]{tran2018spectral}
Tran, B.; Li, J.; and Madry, A. 2018.
\newblock Spectral signatures in backdoor attacks.
\newblock \emph{NeurIPS}, 8000--8010.

\bibitem[{Tsai et~al.(2021)Tsai, Hsu, Yu, and Chen}]{tsai2021formalizing}
Tsai, Y.-L.; Hsu, C.-Y.; Yu, C.-M.; and Chen, P.-Y. 2021.
\newblock Formalizing Generalization and Adversarial Robustness of Neural
  Networks to Weight Perturbations.
\newblock \emph{NeurIPs}, 34.

\bibitem[{Tsai, Chen, and Ho(2020)}]{tsai2020transfer}
Tsai, Y.-Y.; Chen, P.-Y.; and Ho, T.-Y. 2020.
\newblock Transfer learning without knowing: Reprogramming black-box machine
  learning models with scarce data and limited resources.
\newblock \emph{ICML}, 9614--9624.

\bibitem[{Tu et~al.(2019)Tu, Ting, Chen, Liu, Zhang, Yi, Hsieh, and
  Cheng}]{tu2018autozoom}
Tu, C.-C.; Ting, P.; Chen, P.-Y.; Liu, S.; Zhang, H.; Yi, J.; Hsieh, C.-J.; and
  Cheng, S.-M. 2019.
\newblock Autozoom: Autoencoder-based zeroth order optimization method for
  attacking black-box neural networks.
\newblock \emph{AAAI}, 33: 742--749.

\bibitem[{Wang et~al.(2019)Wang, Yao, Shan, Li, Viswanath, Zheng, and
  Zhao}]{wang2019neural}
Wang, B.; Yao, Y.; Shan, S.; Li, H.; Viswanath, B.; Zheng, H.; and Zhao, B.~Y.
  2019.
\newblock Neural cleanse: Identifying and mitigating backdoor attacks in neural
  networks.
\newblock \emph{IEEE S\&P}, 707--723.

\bibitem[{Wang et~al.(2021{\natexlab{a}})Wang, Zhang, Liu, Chen, Xu, Fardad,
  and Li}]{wang2021adversarial}
Wang, J.; Zhang, T.; Liu, S.; Chen, P.-Y.; Xu, J.; Fardad, M.; and Li, B.
  2021{\natexlab{a}}.
\newblock Adversarial attack generation empowered by min-max optimization.
\newblock \emph{NeurIPS}, 34.

\bibitem[{Wang et~al.(2021{\natexlab{b}})Wang, Xu, Liu, Chen, Weng, Gan, and
  Wang}]{wang2021fast}
Wang, R.; Xu, K.; Liu, S.; Chen, P.-Y.; Weng, T.-W.; Gan, C.; and Wang, M.
  2021{\natexlab{b}}.
\newblock On Fast Adversarial Robustness Adaptation in Model-Agnostic
  Meta-Learning.
\newblock \emph{ICLR}.

\bibitem[{Wang et~al.(2020)Wang, Zhang, Liu, Chen, Xiong, and
  Wang}]{wang2020practical}
Wang, R.; Zhang, G.; Liu, S.; Chen, P.-Y.; Xiong, J.; and Wang, M. 2020.
\newblock Practical detection of trojan neural networks: Data-limited and
  data-free cases.
\newblock \emph{ECCV}, 222--238.

\bibitem[{Wang et~al.(2021{\natexlab{c}})Wang, Wang, Chen, Zhao, and
  Lin}]{wangcharacteristi}
Wang, S.; Wang, X.; Chen, P.~Y.; Zhao, P.; and Lin, X. 2021{\natexlab{c}}.
\newblock Characteristic Examples: High-Robustness, Low-Transferability
  Fingerprinting of Neural Networks.
\newblock \emph{IJCAI}.

\bibitem[{Weber et~al.(2020)Weber, Xu, Karlas, Zhang, and Li}]{weber2020rab}
Weber, M.; Xu, X.; Karlas, B.; Zhang, C.; and Li, B. 2020.
\newblock Rab: Provable robustness against backdoor attacks.
\newblock \emph{arXiv preprint arXiv:2003.08904}.

\bibitem[{Weng et~al.(2018{\natexlab{a}})Weng, Zhang, Chen, Song, Hsieh,
  Boning, Dhillon, and Daniel}]{weng2018towards}
Weng, T.-W.; Zhang, H.; Chen, H.; Song, Z.; Hsieh, C.-J.; Boning, D.; Dhillon,
  I.~S.; and Daniel, L. 2018{\natexlab{a}}.
\newblock Towards Fast Computation of Certified Robustness for ReLU Networks.
\newblock \emph{International Coference on ICML}.

\bibitem[{Weng et~al.(2018{\natexlab{b}})Weng, Zhang, Chen, Yi, Su, Gao, Hsieh,
  and Daniel}]{weng2018evaluating}
Weng, T.-W.; Zhang, H.; Chen, P.-Y.; Yi, J.; Su, D.; Gao, Y.; Hsieh, C.-J.; and
  Daniel, L. 2018{\natexlab{b}}.
\newblock Evaluating the Robustness of Neural Networks: An Extreme Value Theory
  Approach.
\newblock \emph{ICLR}.

\bibitem[{Weng et~al.(2020)Weng, Zhao, Liu, Chen, Lin, and
  Daniel}]{weng2020towards}
Weng, T.-W.; Zhao, P.; Liu, S.; Chen, P.-Y.; Lin, X.; and Daniel, L. 2020.
\newblock Towards Certificated Model Robustness Against Weight Perturbations.
\newblock \emph{AAAI}, 6356--6363.

\bibitem[{Wong and Kolter(2018)}]{wong2018provable}
Wong, E.; and Kolter, Z. 2018.
\newblock Provable defenses against adversarial examples via the convex outer
  adversarial polytope.
\newblock \emph{ICML}.

\bibitem[{Wong et~al.(2018)Wong, Schmidt, Metzen, and Kolter}]{wong2018scaling}
Wong, E.; Schmidt, F.~R.; Metzen, J.~H.; and Kolter, J.~Z. 2018.
\newblock Scaling provable adversarial defenses.
\newblock \emph{NeurIPS}.

\bibitem[{Xie et~al.(2020)Xie, Huang, Chen, and Li}]{xie2020dba}
Xie, C.; Huang, K.; Chen, P.-Y.; and Li, B. 2020.
\newblock {DBA}: Distributed Backdoor Attacks against Federated Learning.
\newblock \emph{ICLR}.

\bibitem[{Xu et~al.(2019{\natexlab{a}})Xu, Chen, Liu, Chen, Weng, Hong, and
  Lin}]{xu2019topology}
Xu, K.; Chen, H.; Liu, S.; Chen, P.-Y.; Weng, T.-W.; Hong, M.; and Lin, X.
  2019{\natexlab{a}}.
\newblock Topology attack and defense for graph neural networks: An
  optimization perspective.
\newblock \emph{IJCAI}.

\bibitem[{Xu et~al.(2019{\natexlab{b}})Xu, Liu, Zhao, Chen, Zhang, Fan,
  Erdogmus, Wang, and Lin}]{xu2018structured}
Xu, K.; Liu, S.; Zhao, P.; Chen, P.-Y.; Zhang, H.; Fan, Q.; Erdogmus, D.; Wang,
  Y.; and Lin, X. 2019{\natexlab{b}}.
\newblock Structured adversarial attack: Towards general implementation and
  better interpretability.
\newblock \emph{ICLR}.

\bibitem[{Xu et~al.(2020{\natexlab{a}})Xu, Shi, Zhang, Huang, Chang, Kailkhura,
  Lin, and Hsieh}]{xu2020automatic}
Xu, K.; Shi, Z.; Zhang, H.; Huang, M.; Chang, K.-W.; Kailkhura, B.; Lin, X.;
  and Hsieh, C.-J. 2020{\natexlab{a}}.
\newblock Automatic perturbation analysis on general computational graphs.
\newblock \emph{NeurIPS}.

\bibitem[{Xu et~al.(2020{\natexlab{b}})Xu, Zhang, Liu, Fan, Sun, Chen, Chen,
  Wang, and Lin}]{xu2020adversarial}
Xu, K.; Zhang, G.; Liu, S.; Fan, Q.; Sun, M.; Chen, H.; Chen, P.-Y.; Wang, Y.;
  and Lin, X. 2020{\natexlab{b}}.
\newblock Adversarial t-shirt! evading person detectors in a physical world.
\newblock \emph{ECCV}.

\bibitem[{Yang, Tsai, and Chen(2021)}]{yang2021voice2series}
Yang, C.-H.~H.; Tsai, Y.-Y.; and Chen, P.-Y. 2021.
\newblock Voice2Series: Reprogramming Acoustic Models for Time Series
  Classification.
\newblock In \emph{ICML}.

\bibitem[{Yang et~al.(2019)Yang, Li, Chen, and Song}]{yang2018characterizing}
Yang, Z.; Li, B.; Chen, P.-Y.; and Song, D. 2019.
\newblock Characterizing Audio Adversarial Examples Using Temporal Dependency.
\newblock \emph{ICLR}.

\bibitem[{Zawad et~al.(2021)Zawad, Ali, Chen, Anwar, Zhou, Baracaldo, Tian, and
  Yan}]{zawad2021curse}
Zawad, S.; Ali, A.; Chen, P.-Y.; Anwar, A.; Zhou, Y.; Baracaldo, N.; Tian, Y.;
  and Yan, F. 2021.
\newblock Curse or Redemption? How Data Heterogeneity Affects the Robustness of
  Federated Learning.
\newblock \emph{AAAI}.

\bibitem[{Zhang et~al.(2017)Zhang, Bengio, Hardt, Recht, and
  Vinyals}]{zhang2016understanding}
Zhang, C.; Bengio, S.; Hardt, M.; Recht, B.; and Vinyals, O. 2017.
\newblock Understanding deep learning requires rethinking generalization.
\newblock \emph{ICLR}.

\bibitem[{Zhang et~al.(2022{\natexlab{a}})Zhang, Lu, Zhang, Chen, Chen, Fan,
  Martie, Horesh, Hong, and Liu}]{zhang2022distributed}
Zhang, G.; Lu, S.; Zhang, Y.; Chen, X.; Chen, P.-Y.; Fan, Q.; Martie, L.;
  Horesh, L.; Hong, M.; and Liu, S. 2022{\natexlab{a}}.
\newblock Distributed adversarial training to robustify deep neural networks at
  scale.
\newblock In \emph{Uncertainty in Artificial Intelligence}, 2353--2363. PMLR.

\bibitem[{Zhang et~al.(2018)Zhang, Weng, Chen, Hsieh, and
  Daniel}]{zhang2018efficient}
Zhang, H.; Weng, T.-W.; Chen, P.-Y.; Hsieh, C.-J.; and Daniel, L. 2018.
\newblock Efficient neural network robustness certification with general
  activation functions.
\newblock \emph{NeurIPS}, 4944--4953.

\bibitem[{Zhang et~al.(2019)Zhang, Yu, Jiao, Xing, El~Ghaoui, and
  Jordan}]{zhang2019theoretically}
Zhang, H.; Yu, Y.; Jiao, J.; Xing, E.; El~Ghaoui, L.; and Jordan, M. 2019.
\newblock Theoretically principled trade-off between robustness and accuracy.
\newblock \emph{ICML}, 7472--7482.

\bibitem[{Zhang et~al.(2022{\natexlab{b}})Zhang, Yao, Jia, Yi, Hong, Chang, and
  Liu}]{zhang2022how}
Zhang, Y.; Yao, Y.; Jia, J.; Yi, J.; Hong, M.; Chang, S.; and Liu, S.
  2022{\natexlab{b}}.
\newblock How to Robustify Black-Box {ML} Models? A Zeroth-Order Optimization
  Perspective.
\newblock In \emph{International Conference on Learning Representations}.

\bibitem[{Zhao et~al.(2020{\natexlab{a}})Zhao, Chen, Das, Ramamurthy, and
  Lin}]{Zhao2020Bridging}
Zhao, P.; Chen, P.-Y.; Das, P.; Ramamurthy, K.~N.; and Lin, X.
  2020{\natexlab{a}}.
\newblock Bridging Mode Connectivity in Loss Landscapes and Adversarial
  Robustness.
\newblock \emph{ICLR}.

\bibitem[{Zhao et~al.(2020{\natexlab{b}})Zhao, Chen, Wang, and
  Lin}]{zhao2020towards}
Zhao, P.; Chen, P.-Y.; Wang, S.; and Lin, X. 2020{\natexlab{b}}.
\newblock Towards Query-Efficient Black-Box Adversary with Zeroth-Order Natural
  Gradient Descent.
\newblock \emph{AAAI}.

\bibitem[{Zhao et~al.(2019)Zhao, Liu, Chen, Hoang, Xu, Kailkhura, and
  Lin}]{zhao2019design}
Zhao, P.; Liu, S.; Chen, P.-Y.; Hoang, N.; Xu, K.; Kailkhura, B.; and Lin, X.
  2019.
\newblock On the Design of Black-box Adversarial Examples by Leveraging
  Gradient-free Optimization and Operator Splitting Method.
\newblock \emph{ICCV}, 121--130.

\bibitem[{Zhu et~al.(2019)Zhu, Huang, Li, Taylor, Studer, and
  Goldstein}]{zhu2019transferable}
Zhu, C.; Huang, W.~R.; Li, H.; Taylor, G.; Studer, C.; and Goldstein, T. 2019.
\newblock Transferable clean-label poisoning attacks on deep neural nets.
\newblock \emph{ICML}, 7614--7623.

\end{thebibliography}

\end{document}